\documentclass[lettersize,journal]{IEEEtran}
\usepackage{cite}
\usepackage{amsmath,amssymb,amsfonts}
\usepackage{graphicx,color}
\usepackage{textcomp}
\usepackage{xcolor}
\usepackage{algorithm}
\def\BibTeX{{\rm B\kern-.05em{\sc i\kern-.025em b}\kern-.08em
    T\kern-.1667em\lower.7ex\hbox{E}\kern-.125emX}}
\AtBeginDocument{\definecolor{tmlcncolor}{cmyk}{0.93,0.59,0.15,0.02}\definecolor{NavyBlue}{RGB}{0,86,125}}

\usepackage{comment}
\usepackage{epsfig} % for postscript graphics files
\usepackage{mathptmx} % assumes new font selection scheme installed
\usepackage{times} % assumes new font selection scheme installed
\usepackage{amsmath,amssymb,amsfonts}
\usepackage{pgfplots}
\pgfplotsset{compat=1.7}
\usepackage{pgfplotstable}
\usepackage{tikz,pgfplots}
\usepackage{physics} 
\usepackage{siunitx} 
\usepackage{enumerate} 
\usepackage{verbatim}
\usepackage{color}
\usepackage{mathtools}
\usepackage{gensymb}
\usepackage{breqn}
\usepackage{array}
\usepackage{makecell}
\usepackage{soul}  % we can delete this package later
\usepackage{comment}
\usepackage{subcaption}
\usepackage{algpseudocode} 
\usepackage{multirow}
\usepackage{multicol}
\usepackage{tabularx}
\usepackage{esvect}
\usepackage[table]{xcolor}
\usepackage{placeins} % Add this in the preamble
\usepackage{nameref}
\usepackage[hidelinks]{hyperref}

\begin{document}

\title{Gait-Level Motion Design and Evaluation Framework \\ for Grasp-Based Dynamic Locomotion in Microgravity
}

% \author{Anonymous Authors}

\author{Chaerim Moon, Joohyung Kim, and Justin K. Yim
\thanks{The authors are with the Department of Mechanical Science and Engineering at the University of Illinois at Urbana-Champaign, USA. {\tt\small \{cm74, joohyung, jkyim\}@illinois.edu}}
}



\maketitle
\begin{abstract}

Locomotion in microgravity often relies on sparsely and irregularly arranged anchors, motivating grasp-based mobility with multiple limbs. In this setting, dynamic traversal requires consecutive anchored interactions under coupled dynamic and kinematic constraints, yet the effects of gait-level motion design on locomotion feasibility and performance remain insufficiently understood. This paper formulates the feasibility and performance objectives for grasp-based dynamic locomotion and develops a gait-level parameter-metric framework that relates motion parameters to corresponding evaluation metrics. A physics-based simulation study instantiates the framework across two quadruped morphologies in randomized three-dimensional anchor environments. Controlled variations in gait-level parameters reveal broadly consistent effects on contact support, motion-induced loading, kinematic feasibility, actuation demand, and traversal time across the two robot realizations. These findings suggest that the investigated gait-level parameters provide an interpretable basis for analyzing feasibility and performance trade-offs.

\end{abstract}

% abstract word count limit of 200


\section{Introduction}

\IEEEPARstart{L}{ocomotion} in microgravity is an essential capability in a broad range of space operations. For example, astronauts translate along existing handrails to inspect and maintain the exterior of the International Space Station (ISS) \cite{korona2016extravehicular, moore201021st}. Robonaut2, a humanoid robot, has been developed to assist such tasks, with the dexterity to operate structures and tools designed for humans \cite{diftler2012robonaut, diftler2011robonaut,rehnmark2004experimental}. Another application domain is space construction, where robots traverse sparse truss elements and assemble modular components to build in-orbit infrastructure \cite{jenett2017bill, staritz2001skyworker, spino2025proprioceptive}. Beyond these man-made environments, locomotion in natural settings, such as on the surfaces of asteroids for space mining \cite{yoshida2002novel, candalot2024soft}, offers additional opportunities and challenges for future space exploration.

% \begin{figure}
%     \centering
%     \begin{subfigure}{0.85\columnwidth}
%         \includegraphics[width=0.99\columnwidth]{fig/models/concept_iss.png}
%         \caption{Astronaut transport around the ISS}
%     \end{subfigure}
%     \vspace{3mm}
    
%     \begin{subfigure}{0.85\columnwidth}
%         \includegraphics[width=0.99\columnwidth]{fig/models/concept_asteroid.png}
%         \caption{Asteroid sample transfer}
%     \end{subfigure}
%     \caption{Illustrative scenarios of grasp-based locomotion in low-gravity environments. (a) The ISS 3D models were obtained from \cite{nasa3dmodels}, and (b) the background image of Ryugu is credited to JAXA, University of Aizu, and collaborators.}
    
%     \label{fig:main}
%     \vspace{-5mm}
% \end{figure}

Across both artificial and natural microgravity environments, feasible contact anchors are often sparsely distributed in 3D space, and in natural settings, they may be irregularly arranged. Such conditions challenge locomotion strategies that rely on continuous surface contact, such as wheeled locomotion, and motivate mobility through discrete anchored interactions. Furthermore, microgravity changes the mechanism by which contact contributes to locomotion stability. Unlike terrestrial settings, where gravity provides normal loading that enables frictional support, microgravity requires the robot to actively maintain anchored contact. Consequently, grasp-based locomotion emerges as an effective and practical mode of mobility in these environments.

% Such conditions inherently challenge locomotion strategies that rely on continuous ground contact, making limbed locomotors more viable than wheeled counterparts. However, microgravity introduces additional requirements compared to terrestrial locomotion, particularly in how a robot needs to regulate its interaction forces to achieve controlled and reliable motion. Unlike terrestrial settings, where gravity-induced friction readily stabilizes motion, microgravity necessitates deliberate, anchored interactions. Consequently, grasp-based locomotion emerges as an effective and practical mode of mobility in these environments.

\begin{figure}
    \centering
    \includegraphics[width=0.85\columnwidth]{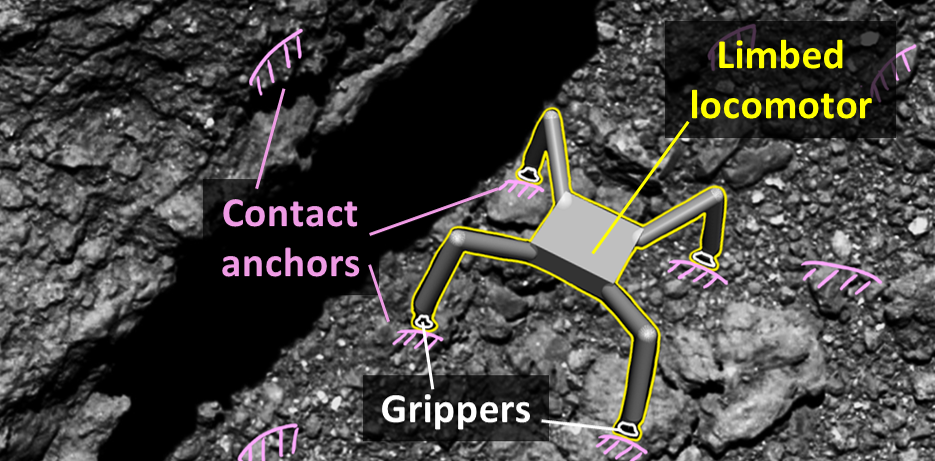}
    \caption{Illustrative scenario of grasp-based locomotion in low-gravity environments. The background image of the asteroid Ryugu is credited to JAXA, University of Aizu, and collaborators.}
    \label{fig:main}
    \vspace{-5mm}
\end{figure}

Despite its potential influence on space exploration, dynamic locomotion in microgravity remains insufficiently investigated. Many existing approaches adopt quasi-static assumptions, treating motion-induced inertial effects as negligible \cite{chen2024locomotion, zhang2013capuchin, bretl2006motion}. While effective for extremely cautious maneuvers, these approaches offer limited guidance for time-critical or energy-limited mission operations. Other efforts relax the quasi-static assumption and account for whole-body dynamics, for example by coordinating base and limb motions to regulate motion-induced reactions and avoid contact loss \cite{ribeiro2023ramp}. However, this strategy still imposes unnecessary restrictions on the motion design space by restricting base translation to a separate phase after limb swinging. This motivates the development of a framework to support motion design under the constraints specific to dynamic locomotion in microgravity.

% This motivates a systematic characterization of gait-level motion parameters and corresponding evaluation criteria for dynamic grasp-based locomotion in microgravity.

% These limitations call for a systematic investigation into dynamic locomotion in microgravity.

% This paper presents a gait-level framework for designing and evaluating grasp-based dynamic locomotion in microgravity, motivated by the scenarios illustrated in Fig. \ref{fig:main}. The considered locomotion scenarios require full 6D limb manipulation to interact with the available contact anchors. The main contributions of this paper are summarized as follows:

This paper addresses grasp-based dynamic locomotion in microgravity from a gait-level perspective. The considered locomotion involves full 6D limb manipulation across irregularly distributed contact anchors, as illustrated by the scenario in Fig. \ref{fig:main}, and requires motion design to account for coupled feasibility and performance considerations. The main contributions of this paper are summarized as follows:

% This paper investigates dynamic grasp-based locomotion in microgravity across irregularly distributed contact anchors, as illustrated by the scenario in Fig. \ref{fig:main}. The considered locomotion involves full 6D limb manipulation, requiring motion design to account for coupled feasibility and performance considerations. The main contributions of this paper are summarized as follows:

\begin{itemize}
    \item Formulation of feasibility and performance objectives for grasp-based dynamic locomotion in microgravity 
    \item Development of a gait-level parameter–metric framework for systematically relating motion-design choices to feasibility and performance outcomes
    \item Physics-based realization and comparative evaluation across randomized environments and representative quadruped morphologies
\end{itemize}

\section{Related Work}

\subsection{Grasp-Based Locomotion}

Previous grasp-based locomotion studies provide useful insights into contact and whole-body motion requirements. Under Earth's gravity, locomotion through grasping interactions with environmental structures (e.g., bouldering holds, poles, or walls) has been explored, especially in the context of climbing robots. In structured environments, traversal can be achieved by alternating hold-and-release sequences using grippers \cite{tanaka2022scaler, uno2021hubrobo}. Other approaches generate pushing forces against surrounding surfaces to exploit friction for support and propulsion \cite{lin2019optimization, zhang2021transition}. In both cases, reliable contact transitions are essential and often explored using sampling-based methods \cite{bretl2006motion, zhang2013capuchin, kingston2022scaling}; however, dynamic locomotion involving non-negligible motion-induced inertial effects remains relatively underexplored. Grasp-based locomotion concepts have also been studied in space robotics, mainly along two complementary directions. One is to regulate motion-induced inertial effects by designing swing motions that reduce momentum variation and adjusting the base pose to compensate for the resulting whole-body momentum change \cite{ribeiro2023ramp, ribeiro2023mobility}. The other is to enlarge the admissible grasping range through gripper design, including electrostatic adhesion \cite{xiang2024design} and microspine-based anchoring \cite{parness2011anchoring}.

Although these studies establish important mechanisms for grasp-based traversal, they primarily focus on contact transition planning, momentum regulation, or gripper capability. Less attention has been given to how the motion design problem of grasp-based dynamic locomotion should be formulated under coupled feasibility and performance considerations.

% Here, dynamic locomotion denotes locomotion involving non-negligible motion-induced inertial wrenches.

% Although previous studies have established important groundwork for grasp-based locomotion, including in low-gravity settings, gaps remain in the investigation of generalizable feasibility conditions and the exploration of gait parameters.

\subsection{Gait-Level Descriptions of Locomotion}

% To study gait-level effects, terrestrial locomotion research has relied on gait-level gait descriptors to characterize whole-body behavior. Hildebrand gait diagrams represent the contact timing structure of a gait through swing sequence, duty factor, and phase offsets \cite{}. Stride length and locomotion speed characterize the spatial progression and temporal execution of a gait, respectively \cite{}. The center of mass (COM) height has also been considered, as it defines the dynamics of pendulum-based locomotion templates \cite{}.

Gait-level descriptions have been introduced to characterize terrestrial locomotion. Gait analysis provides an abstract representation of contact schedules through quantities such as footfall sequence, duty factor, relative phase, stride length, and locomotion speed \cite{hildebrand1965symmetrical,alexander1984gaits}. Reduced-order models such as the linear inverted pendulum further characterize locomotion dynamics using gait-level quantities, including center-of-mass (COM) height \cite{kajita20013d}. Such descriptions have subsequently been adopted in robotic gait generation to parameterize contact organization, interlimb coordination, locomotion progression, and body posture \cite{margolis2023walk, watanabe2025hierarchical, iscen2018policies}.

These descriptions provide a basis for locomotion design across multi-limbed systems. In grasp-based locomotion in microgravity, gait-level choices are likewise expected to shape how locomotion is organized and executed. However, how these choices affect locomotion through the underlying mechanisms remains insufficiently understood. Assessing these effects requires quantitative measures that capture the resulting feasibility and performance outcomes.

% Gait-level choices can alter the active grasp configuration, progression between anchors, and body motion relative to the available contacts, with consequences for both locomotion feasibility and performance.

% These descriptions provide a basis for locomotion design across multi-limbed systems. However, their implications for grasp-based locomotion in microgravity remain insufficiently understood. They are expected to influence such as active grasp configurations, anchor-to-anchor progression, traversal duration, and body motion relative to available anchors. Assessing these effects requires feasibility and performance criteria appropriate to grasp-based locomotion in microgravity.

\begin{figure}
    \centering
    \begin{subfigure}{0.57\columnwidth}
        \includegraphics[width=0.99\columnwidth]{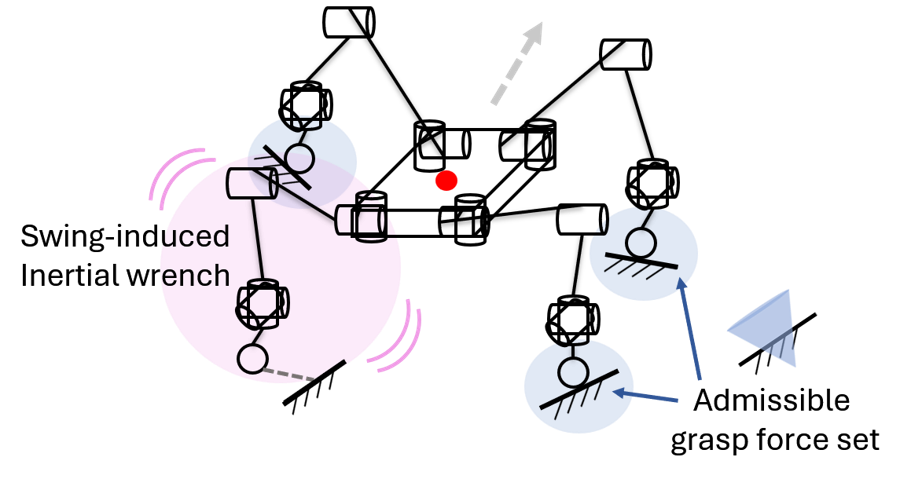}
        \caption{Contact and motion wrenches}
        \label{fig:whole_body_wrench}
    \end{subfigure}
    \begin{subfigure}{0.38\columnwidth}
        \includegraphics[width=0.99\columnwidth]{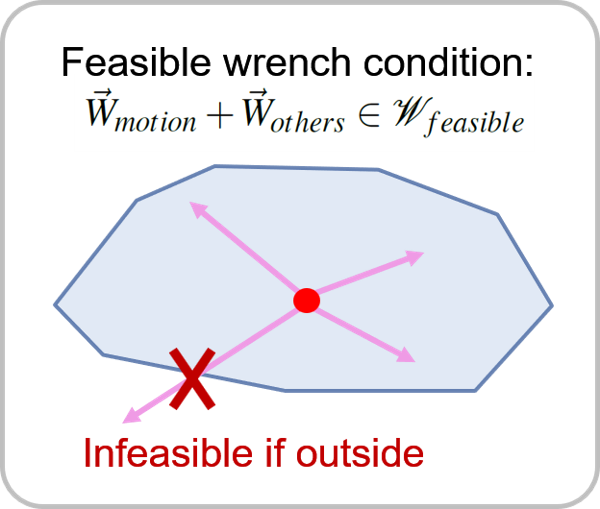}
        \caption{Feasibility condition}
        \label{fig:wrench_feasibility_viz}
    \end{subfigure}
    \caption{Dynamic feasibility in grasp-based locomotion. (a) Active grasps define contact-wrench capacity, while robot motion generates inertial wrench demand. (b) The applied wrench must remain within the feasible wrench set.}
    \vspace{-3mm}
    \label{fig:wrench_feasibility}
\end{figure}

\section{Feasibility and Performance Objectives}

The feasibility requirements and performance objectives associated with the target scenario are identified to clarify what motion designs must satisfy and which aspects of locomotion should be improved. These objectives establish the basis for characterizing gait-level motion designs.

% The target scenario involves traversal across randomly distributed contact anchors. The feasibility requirements and performance objectives associated with this scenario are first identified, providing the basis for selecting the gait-level parameters and evaluation metrics.

% In the targeted scenario of grasp-based locomotion with full 6D limb manipulation, locomotion feasibility is governed by tightly coupled dynamic and kinematic considerations. This section defines the required fundamental feasibility constraints.

\subsection{Dynamic Feasibility}
The criteria for assessing dynamic stability differ from those in terrestrial settings. In terrestrial locomotion, locomotion stability is often approximated using the support polygon \cite{bellicoso2017dynamic, pongas2007robust, kalakrishnan2011learning}. In contrast, such simplifications no longer hold for grasp-based locomotion in microgravity, where key assumptions, including unilateral, compressive ground contact, are violated. Thus, a comprehensive analysis of the feasible 6D wrench space is required. 

% It defines the range of whole-body accelerations that can be achieved without contact detachment, which is critical in microgravity, where contact loss may lead to uncontrolled drift and subsequent mission failure.

% The feasible net contact wrench space can be represented as a wrench polytope, denoted as $\mathcal{W}_{feasible} \subset \mathbb{R}^6$, whose element is defined as $\vec{W}$=$[\mathbf{f}^\top\ \boldsymbol{\tau}^\top]^\top$.

Dynamic feasibility is determined by whether the active grasp configuration can support the wrench generated by robot motion and external disturbances (Fig. \ref{fig:wrench_feasibility}). This criterion is particularly critical in microgravity, as contact loss may lead to uncontrolled drift and subsequent mission failure. The set of feasible net contact wrenches can be represented as a wrench polytope in a common reference frame, such as the base frame. This set is denoted by $\mathcal{W}_{feasible} \subset \mathbb{R}^6$, whose element is defined as $\vec{W}$=$[\mathbf{f}^\top\ \boldsymbol{\tau}^\top]^\top$.
It depends on parameters including grasp geometry, contact locations, allowable force directions and magnitudes, and friction constraints. The applied net wrench arises from various sources. Specifically, $\vec{W}_{motion}$ denotes the wrench generated by the robot's own motion-induced inertial effects, whereas $\vec{W}_{others}$ captures additional wrenches arising from the unpredictable inertial effects associated with carried payloads or incidental collisions. Contact stability during locomotion requires this net wrench to remain within the feasible wrench space:

\begin{equation}
    \vec{W}_{motion} + \vec{W}_{others} \in \mathcal{W}_{feasible}
    \label{eq:wrench_cond}
\end{equation}

This condition highlights two complementary ways to improve robustness. One is to enlarge the feasible wrench space $\mathcal{W}_{feasible}$ through favorable contact configuration selection. The other is to limit the motion-induced wrench $\vec{W}_{motion}$ through coordination of base and limb motions.

\subsection{Kinematic Feasibility}

Grasp-based locomotion relies on consecutive anchor-to-anchor transitions, making kinematic feasibility an additional primary constraint. In 3D environments with irregularly placed feasible contact anchors, the system must reach and engage the scheduled anchors throughout the gait cycle. As a quadruped includes a floating base, base motion strongly affects this feasibility by shaping each limb's attainable contact pose set in the world frame. It highlights the importance of regulating the relative geometry between the robot and its surrounding structures through base trajectory planning.

\subsection{Transportation Efficiency}

Transportation efficiency is also an important aspect for space robotic systems, as they often operate under constrained power and energy resources. High actuation demand reduces the resource margin for subsequent operations, whereas long traversal duration delays mission tasks such as inspection, construction, and sample transfer. More critically, onboard energy depletion over excessively long operating durations may compromise continued locomotion and grasp retention, potentially resulting in the loss of the robot. Therefore, among motions that satisfy dynamic and kinematic feasibility, locomotion behaviors that reduce the demand and time required to complete the traversal are preferred.

\section{Motion Parameters and Evaluation Metrics}

% Based on the feasibility constraints and performance objectives defined in the previous section, this section identifies a compact set of gait-level parameters that influence contact wrench support, motion-induced loading, kinematic feasibility, and transportation efficiency, and defines the metrics used to evaluate their effects.

% The proposed gait motion design framework is formulated with consideration of the feasibility constraints and performance objectives. Five selected parameters characterize the gait-level organization and execution of locomotion, while complementary metrics evaluate the resulting feasibility and performance.

% To investigate how locomotion design affects feasibility and performance, motion parameters and evaluation metrics are defined. The motion parameters enable controlled variations in locomotion, while the evaluation metrics characterize the resulting feasibility and performance.

Feasibility and performance objectives define the requirements of locomotion, but do not directly prescribe how the motion should be organized and executed. Bridging this gap requires understanding how gait-level design decisions affect these objectives and whether improvements in one objective are accompanied by benefits or compromises in others. To support such an analysis, a gait-level motion parameterization and quantitative evaluation metrics are introduced.

\subsection{Motion Parameters}

% Motion parameters that form a compact set of gait-level variables spanning contact sequencing, interlimb timing, spatial progression, motion execution rate, and body-to-contact geometry are selected. They are swing order, swing-phase overlap, stride length, base travel speed, and base height. Unlike joint-level trajectory or controller parameters, these variables are defined upstream of a specific whole-body realization and remain meaningful across different multi-limbed morphologies.

% They are not intended to exhaustively describe whole-body motion; they form a compact set of gait-level variables spanning contact sequencing, interlimb timing, spatial progression, motion execution rate, and body-to-contact geometry. 

The gait-level parameterization consists of a compact set of variables that capture contact organization, spatial and temporal progression, and base–anchor geometry without being tied to a specific robot morphology or whole-body realization. Accordingly, five parameters are selected: swing order, swing-phase overlap, stride length, swing speed, and base height. In the considered scenarios, swing order determines which limb is released at each contact transition, while swing-phase overlap describes the temporal overlap between consecutive limb swings. Stride length specifies the nominal spatial progression between successive anchor configurations, whereas swing speed determines the rate of swing execution. Base height is defined as the nominal offset of the base from the local anchor plane.

\subsection{Evaluation Metrics}
% The research questions require evaluation metrics that quantify support capability, motion-induced disturbance, actuation demand, and traversal time. Kinematic feasibility is not included among the quantitative performance metrics, as it is treated as a hard constraint to be rejected if infeasible.

% The research questions require evaluation measures that quantify how the selected gait-level parameters relate to the defined feasibility and performance objectives. 

The evaluation metrics address dynamic feasibility and transportation efficiency. Kinematic feasibility is instead assessed as a binary outcome based on whether joint trajectories exist that realize the prescribed base and end-effector motions throughout the gait cycle while respecting joint limits. The factors governing dynamic feasibility are characterized using support capability and motion-induced disturbance, representing the contact-side capacity and motion-side demand, respectively. Transportation efficiency is evaluated using actuation demand and traversal time.

\subsubsection{Support Capability}
% Note that the contact-set score and the support capability metric serve distinct purposes. The former is a geometry-based predictor used to rank candidate swing orders; the latter evaluates the support provided by the realized whole-body configuration. 

Support capability is quantified in terms of disturbance-rejection capacity, since the 6D feasible wrench polytope does not directly provide a scalar measure for comparing different grasp configurations. This capacity can be evaluated using the maximum external point force that can be balanced by the active grasp configuration. For a set of points $\mathcal{P}$ on the robot body, consider a point force of magnitude $\lambda$ applied at $\mathbf{p}\in\mathcal{P}$ along a unit direction $\mathbf{u}\in\mathcal{U}_{\mathbf{p}}$, where $\mathcal{U}_{\mathbf{p}}$ denotes the set of disturbance directions at $\mathbf{p}$. For an active contact configuration $\mathcal{C}$, the maximum supportable force in this direction is defined as

\begin{equation}
F(\mathcal{C},\mathbf{p},\mathbf{u})
:=
\max_{\lambda \geq 0}
\left\{
\lambda
\;\middle|\;
-\lambda
\begin{bmatrix}
\mathbf{u} \\
\mathbf{r}_{\mathbf{B}\mathbf{p}}\times\mathbf{u}
\end{bmatrix}
\in
\mathcal{W}_{\mathrm{feasible}}^{\mathbf{B}}(\mathcal{C})
\right\},
\label{eq:directional_poking_capacity}
\end{equation}

where $\mathbf{r}_{\mathbf{B}\mathbf{p}}$ denotes the position vector from the origin of the realized base frame $\mathbf{B}$ to $\mathbf{p}$, and $\mathcal{W}_{\mathrm{feasible}}^{\mathbf{B}}(\mathcal{C})$ is the feasible contact wrench set expressed about that frame. 

The support capability metric is obtained by taking the minimum disturbance-rejection capacity over the body points and disturbance directions:

\begin{equation}
F_{support}(\mathcal{C})
:=
\min_{\mathbf{p} \in \mathcal{P}}
\min_{\mathbf{u}\in\mathcal{U}_{\mathbf{p}}}
F(\mathcal{C},\mathbf{p},\mathbf{u}).
\label{eq:minimum_disturbance_resistance}
\end{equation}

The resulting metric represents the largest point-force magnitude that can be balanced for all $\mathbf{p}$ and $\mathbf{u}$.

\subsubsection{Motion-induced Disturbance}
Motion-induced disturbance is evaluated using two wrench-based metrics: whole-body motion-induced wrench and swing-induced wrench. The whole-body motion-induced wrench quantifies the net momentum variation induced by the planned gait motion under grasp support. The swing-induced wrench quantifies the inertial wrench generated specifically by the swinging limbs; it measures the wrench transferred from the swinging limbs to the base frame, aggregated over all limbs in swing.

\subsubsection{Actuation Demand}
Actuation demand is captured by the total mechanical work:
\begin{equation}
W \;=\; \int_{t_0}^{t_f} \sum_{j} \left|\tau_j(t)\,\omega_j(t)\right|\, dt,
\end{equation}
where $\tau_j(t)$ and $\omega_j(t)$ denote the actuating torque and angular velocity of joint $j$, respectively. Alongside total work, actuation demand can be further characterized using the joint torque and velocity.

\subsubsection{Traversal Time}
Traversal time is defined as the elapsed time required to complete a traversal task under the specified gait parameter constraints.

% \begin{figure*}
%     \centering
%     \includegraphics[width=0.63\linewidth]{fig/motion_planning/planning_architecture.png}
%     \caption{Planning architecture for grasp-based microgravity locomotion with multi-limbed robotic systems. Each layer addresses distinct functions: stride goal planning; coordinated base and limb motion planning; and execution with stability monitoring.
%     }
%     \label{fig:motion_planning_framework}
% \end{figure*}

\section{Simulation}

\begin{figure}
    \centering
    \includegraphics[width=0.99\linewidth]{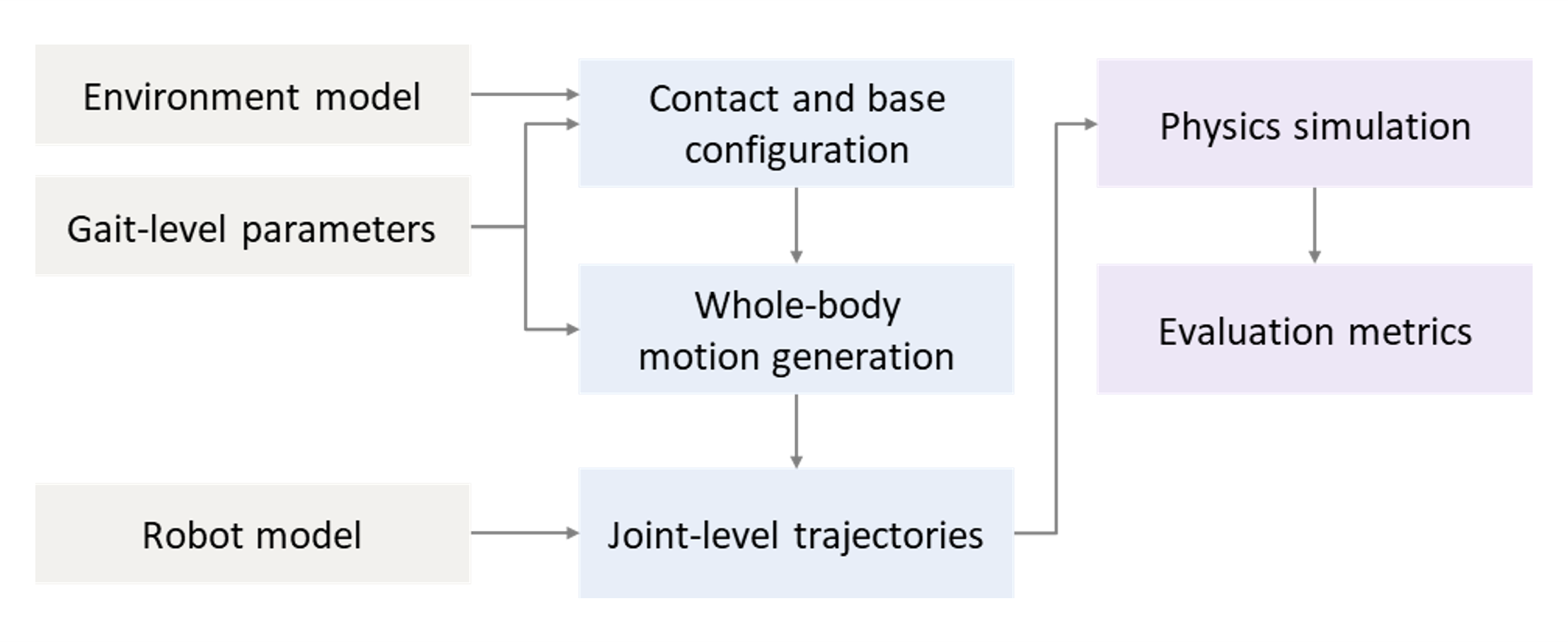}
    \caption{Parameterized motion generation pipeline for instantiating prescribed gait-level parameters as executable motions.
    }
    % \vspace{-3mm}
    \label{fig:motion_planning_framework}
\end{figure}

A physics-based simulation study is conducted in MuJoCo to demonstrate that the proposed gait-level parameters and evaluation metrics can be realized within a complete locomotion framework. The motion generation realization used in this study is summarized in Fig. \ref{fig:motion_planning_framework}. The selected robot models, environment representations, and motion generation approach constitute one specific instantiation rather than a unique or prescriptive implementation. Within this instantiation, controlled variations of the gait-level parameters are used to demonstrate how the proposed framework exposes interpretable relationships between gait-level design choices and locomotion outcomes. All simulations are conducted under idealized microgravity by setting the gravitational acceleration to zero, $\vec{g}=\vec{0}$.

\begin{figure}
    \centering
    \begin{subfigure}{0.40\columnwidth}
        \includegraphics[width=0.85\columnwidth]{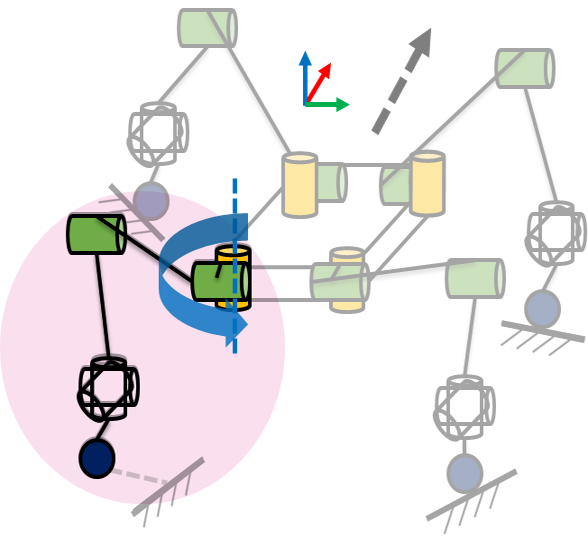}
        \caption{Quadruped \#1}
        \label{fig:quadruped_ypp}
    \end{subfigure}
    \begin{subfigure}{0.40\columnwidth}
        \includegraphics[width=0.85\columnwidth]{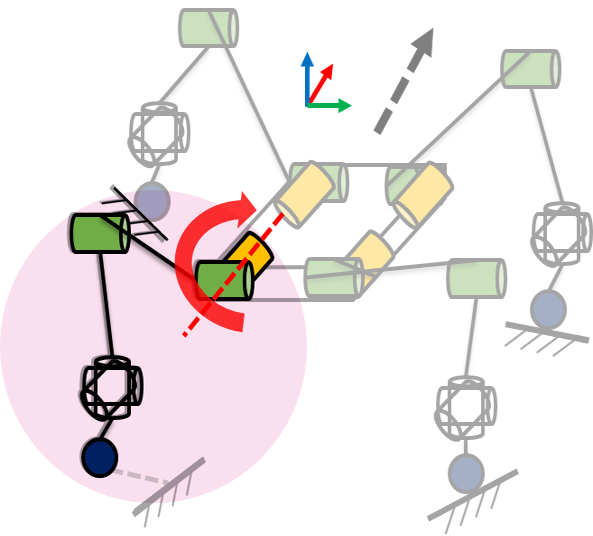}
        \caption{Quadruped \#2}
        \label{fig:quadruped_rpp}
    \end{subfigure}
    \caption{Selected 6-DOF quadruped configurations. The distinction lies in the proximal chain: (a) YPP and (b) RPP.}
    \vspace{-3mm}
    \label{fig:quadruped_morphologies}
\end{figure}

\subsection{Robot Models}

% For tractable evaluation, the framework is instantiated on a quadrupedal system, which represents a minimal morphology that still supports nontrivial contact scheduling and transitions between support configurations with different numbers of active contacts. The robot consists of a rigid base and four identical 6-DOF limbs, each equipped with a gripper capable of generating pulling forces at the end-effector. Each limb comprises 2-DOF hip joints, a 1-DOF knee joint, and a 3-DOF ankle. The robot specifications are summarized in Table \ref{table:robot_model}. The base mass is chosen to dominate the limb masses to reflect heavy-load microgravity operation scenarios, and link masses are assigned in proportion to link lengths, assuming a uniform linear density of 1 kg/m.

Among available multi-limbed robotic systems, a quadrupedal system is selected for evaluation as it represents a minimal morphology that still supports nontrivial contact scheduling and transitions between support configurations with different numbers of active contacts. The robot consists of a rigid base and four identical 6-DOF limbs, each equipped with a gripper capable of generating pulling forces up to 130 N at the end-effector. Each limb comprises a 2-DOF hip, a 1-DOF knee, and a 3-DOF ankle.

% The robot specifications are summarized in Table \ref{table:robot_model}. The base mass is chosen to dominate the limb masses to reflect heavy-load microgravity operation scenarios, and link masses are assigned in proportion to link lengths, assuming a uniform linear density of 1 kg/m.

% In both cases, the distal chain is configured as a spherical joint to align the end-effector orientation with the grasp pose.

Two quadruped morphologies are considered (Fig. \ref{fig:quadruped_morphologies}): Yaw-Pitch-Pitch (YPP) \cite{hooks2020alphred, uno2021hubrobo, leuthard2024magnecko} and Roll-Pitch-Pitch (RPP) \cite{katz2019mini, hutter2016anymal, hutter2012starleth}, which differ in the proximal joint-axis configuration while sharing an identical distal chain. The two morphologies are not compared to identify a preferred mechanical design; rather, they provide distinct kinematic realizations for assessing whether parameter-metric trends persist across robot-specific motion mappings. 

The geometric, inertial, and actuator parameters of the base and proximal legs are adopted from the Unitree B2 robot description \cite{unitree_ros}. A 3-DOF spherical ankle is appended to the distal end of each leg to align the end-effector orientation with the grasp pose. The wrist is modeled as three intersecting revolute joints and parameterized using the Deep Robotics J60-6 actuator specifications. Each limb is controlled using a position-based torque controller to track the planned motions while isolating the effects of gait parameters from controller-level optimization. Multi-contact interactions are handled implicitly by MuJoCo's constrained dynamics solver. Contact reactions under simultaneous grasp constraints are therefore determined by the constrained dynamics rather than through a separately prescribed internal-force distribution.

Joint torques for both swing and stance limbs are given by

\begin{equation}
    \begin{aligned}
    \tau = M(q)\big(\ddot{q}_{des} + K_P e + K_D \dot{e} \big) + C(q,\dot{q})\dot{q}
    \end{aligned}
\end{equation}

where $M(q)$ denotes the joint-space inertia matrix, $C(q,\dot{q})$ is the centrifugal and Coriolis term, and $e = q_{des}-q$ and $\dot e=\dot q_{des}-\dot q$ are the joint position and velocity tracking errors. $K_P$ and $K_D$ are the proportional and derivative gain matrices, respectively. Note that the gravity term is omitted due to the microgravity assumption.

For both robot morphologies, the end-effector and grasp interaction are represented using an idealized bounded-attachment model that permits compressive contact and bounded adhesive pulling. The tangential, torsional, and rolling friction coefficients are set to 1, 0.005, and 0.0001, respectively. In MuJoCo, each end-effector is modeled as a spherical geometry equipped with an adhesion actuator that applies a bounded pulling force along the contact normal. Additional contact geometries are included to prevent unconstrained rolling about the contact normal.

The contact interaction bounds and force limits required for the contact-wrench analysis are adopted from the physics simulation model, ensuring consistency between the simulated grasp interactions and the evaluated wrench capacity. Accordingly, each active grasp is represented as the combination of a unilateral frictional contact force and a bounded adhesive force applied along the negative contact normal. The resulting net contact force set forms a friction cone shifted along the negative contact normal, which is approximated using a polyhedral representation, as shown in Fig. \ref{fig:wrench_feasibility}.

% In simulation, the end-effector geometry is abstracted as a sphere, and microspine grasping is abstracted as a bounded attachment interaction that captures load-limited engagement and detachment. Attachment is implemented using MuJoCo's adhesion actuator with explicit capacity limits, detaching when the load exceeds these limits. To improve contact stability, the gripper incorporates peripheral contact points that distribute attachment loads across multiple contacts, analogous to passive self-alignment in existing gripper designs. Grasping support is realized through modified friction coefficients. For contact wrench space evaluation, the contact model combines a polyhedral friction cone with a spherical, omnidirectional grasp model.

\begin{figure}
    \centering
    \includegraphics[width=0.85\linewidth]{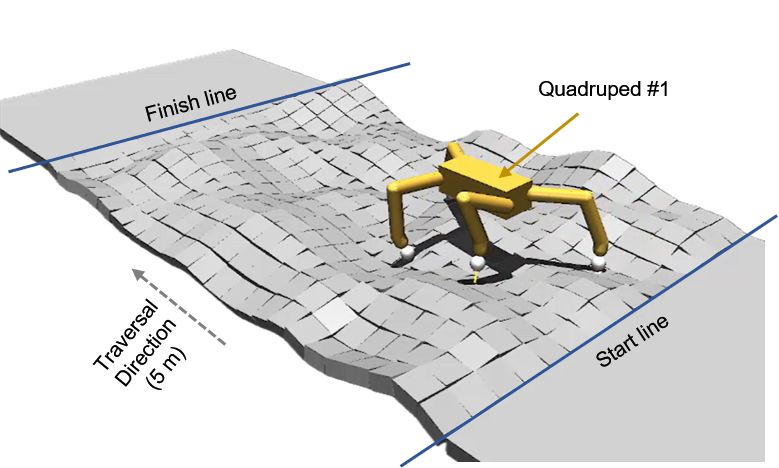}
    \caption{Visualization of the simulation environment}
    \vspace{-5mm}
    \label{fig:simulation}
\end{figure}

\subsection{Environment Models}
To represent irregular grasp-based locomotion environments, randomized three-dimensional terrains are generated for a 5-m forward traversal task. Each terrain consists of patches with varying elevations and surface orientations, and is constructed to include height differences of 0.3 m, as represented in Fig. \ref{fig:simulation}. Patch orientations are bounded within $\pm$22.5$\degree$, with a spatial orientation variation of 15–30$\degree$/m. The orientation difference between adjacent patches is limited to 30$\degree$. Candidate contact poses are defined at the center of each terrain patch, with the end-effector orientation aligned with the corresponding surface normal.

% \begin{figure}
%     \centering
%     \includegraphics[width=0.90\columnwidth]{fig/motion_planning/hildebrand_diagram.png}
%     \caption{Hildebrand representation of the parameterized gait cycle. The swing-phase overlap between the $i_{th}$ and $(i+1)_{th}$ swings is denoted by $\alpha_i$.}
%     \vspace{-5mm}
%     \label{fig:gait_patterns}
% \end{figure}

\subsection{Motion Generation}

% A parameterized locomotion planning pipeline (Fig. \ref{fig:motion_planning_framework}) is used to generate executable whole-body motions under prescribed gait-level parameters. Its role is to provide consistent motion realizations for controlled parameter sweeps and evaluation. Anchor identification and robot-to-environment pose estimation are assumed to be provided by existing vision-based perception and SLAM modules.

% At the stride level, candidate target contacts are selected according to the prescribed stride length, and the nominal target base pose is constructed from the target contact geometry and base height. Swing speed determines the duration of each step, while swing order and swing-phase overlap determine the interlimb coordination of the gait cycle. 

% The gait cycle is visualized using the Hildebrand diagram in Fig. \ref{fig:gait_patterns}. Overlap is permitted only for swing pairs (1--2) and (3--4), while the remaining swing transitions are constrained to zero overlap to avoid prolonged two-limb-support phases. For each permitted pair, the overlap is applied symmetrically and defined relative to the shorter of the two swing durations. The resulting smooth base and end-effector trajectories are then converted to joint-level commands via inverse kinematics.

The motion generation procedure instantiates the prescribed gait-level parameters as executable whole-body motions. Given these parameters, candidate target contacts are selected according to the prescribed stride length, and the nominal target base pose is constructed from the target contact geometry and base height. Anchor identification and robot-to-environment pose estimation are assumed to be provided by existing vision-based perception and SLAM modules. 

% Overlap is permitted only for swing pairs (1--2) and (3--4), while the remaining swing transitions are constrained to zero overlap to avoid prolonged two-limb-support phases, which do not span the full dimensional wrench space.

Temporal coordination is imposed according to the prescribed swing order, swing speed, and swing-phase overlap. Swing overlap is allowed only between every other step, instead of at the transition between all steps, to avoid prolonged two-limb-support phases, which do not span the full dimensional wrench space. For each permitted pair, the overlap is applied symmetrically and defined relative to the shorter of the two swing durations. Based on these spatial and temporal specifications, smooth base and end-effector trajectories are generated and converted to joint-level commands via inverse kinematics.

% The resulting gait cycle is visualized using the Hildebrand diagram in Fig. \ref{fig:gait_patterns}. 

% Smooth base and end-effector trajectories are then generated between the current and target configurations, and inverse kinematics maps these trajectories to joint-level commands for simulation.

% All parameter conditions are instantiated using the same motion-generation and execution procedures. This is intended to ensure that the observed differences primarily reflect the prescribed gait-level variables.

\subsection{Comparative Study Design}

To investigate the gait-level parameters of interest, a reference condition and a set of parameter variants are considered. Each variant changes one parameter while keeping the others at their reference values. The evaluated conditions are intended to provide representative comparisons rather than an exhaustive sampling of the parameter space. This one-at-a-time design is used to isolate interpretable first-order relationships between individual gait-level choices and the corresponding evaluation metrics.

% These values specify the present simulation instantiation without restricting the parameter and metric definitions.

The parameter values are selected within ranges that permit comparable motion generation for both robot morphologies. The reference condition represents a non-extreme setting within these ranges and is not intended as an optimum. The variants are sufficiently separated from the reference values to represent meaningfully distinct contact geometries or motion timings. The reference condition uses the score-ranked swing order, a swing-phase overlap of 30\%, a stride length of 0.2~m, a swing speed of 1.2~rad/s, and a nominal base height of 0.5~m. The tested variants are amble (LR-LF-RR-RF) and trot (LR-RF-RR-LF) for swing order; no swing-phase overlap; 0.4~m for stride length; 1.0~rad/s for maximum swing speed; and 0.4~m for nominal base height. 

% Note that two conditions require additional definition: the score-ranked swing order and the minimum-feasible overlap.

% The score-ranked swing order is determined using a contact-set score that quantifies the induced contact configurations, without requiring whole-body motion realization. The score is evaluated in a nominal base frame $\mathbf{B}_n$, constructed from the target contact poses by aligning it with the best-fit contact plane and offsetting it by the prescribed base height. For an active contact configuration $\mathcal{C}$, let $\mathcal{W}_f^{\mathbf{B}_n}(\mathcal{C})$ and $\mathcal{W}_{\tau}^{\mathbf{B}_n}(\mathcal{C})$ denote the force and torque projections of the feasible contact wrench set. For each disturbance class $\zeta \in \{a,s,b\}$, corresponding to arbitrary, swing-induced, and base-motion-induced disturbances, both the minimum and mean directional scores are considered.

The score-ranked swing order is determined using a contact-set score that quantifies the induced contact configurations without requiring whole-body motion realization. The score is evaluated in a nominal base frame $\mathbf{B}_n$, constructed from the target contact poses by aligning it with the best-fit contact plane and offsetting it by the prescribed base height. For an active contact configuration $\mathcal{C}$, let $\mathcal{W}^{\mathbf{B}_n}(\mathcal{C})$ denote the feasible contact wrench set, with force and torque projections $\mathcal{W}_f^{\mathbf{B}_n}(\mathcal{C})$ and $\mathcal{W}_{\tau}^{\mathbf{B}_n}(\mathcal{C})$, respectively. For each disturbance class $\zeta \in \{a,s,b\}$, corresponding to arbitrary, swing-induced, and base-motion-induced disturbances, both the minimum and mean directional scores are considered.

For the arbitrary disturbance class, force and torque capacities are evaluated separately because no prescribed pairing exists between their directions.
\begin{equation}
\begin{aligned}
\small
s_a(\mathcal{C})
=
\sum_{\kappa\in\{\min,\mathrm{mean}\}} w_\kappa
\Bigg[
&\kappa_{\mathbf{u}\in\mathcal{U}_{a,f}^{\mathbf{B}_n}}
\left(
\max_{\mathbf{f}\in\mathcal{W}_f^{\mathbf{B}_n}(\mathcal{C})}
\mathbf{f}^{\top}\mathbf{u}
\right)
\\
&+
\kappa_{\mathbf{u}\in\mathcal{U}_{a,\tau}^{\mathbf{B}_n}}
\left(
\max_{\boldsymbol{\tau}\in\mathcal{W}_{\tau}^{\mathbf{B}_n}(\mathcal{C})}
\frac{\boldsymbol{\tau}^{\top}\mathbf{u}}{\ell_{\mathrm{nom}}}
\right)
\Bigg],
\end{aligned}
\label{eq:arbitrary_disturbance_score}
\end{equation}
where $\mathcal{U}_{a,f}^{\mathbf{B}_n}$ and $\mathcal{U}_{a,\tau}^{\mathbf{B}_n}$ contain the prescribed force and torque directions, $\kappa$ denotes the minimum or mean operator, and $\ell_{\mathrm{nom}}$ is the nominal distance from the wrench reference point to the active contacts.

For the swing-induced and base-motion-induced disturbance classes
$\zeta\in\{s,b\}$, the force and torque components are evaluated jointly
using the same feasible wrench:
\begin{equation}
\small
s_\zeta(\mathcal{C})
=
\sum_{\kappa\in\{\min,\mathrm{mean}\}} w_\kappa\,
\kappa_{(\mathbf{u}_f,\mathbf{u}_\tau)\in\mathcal{U}_{\zeta}^{\mathbf{B}_n}}
\max_{(\mathbf{f},\boldsymbol{\tau})\in\mathcal{W}^{\mathbf{B}_n}(\mathcal{C})}
\left(
\mathbf{f}^{\top}\mathbf{u}_f
+
\frac{\boldsymbol{\tau}^{\top}\mathbf{u}_\tau}{\ell_{\mathrm{nom}}}
\right),
\label{eq:motion_disturbance_score}
\end{equation}
where $\mathcal{U}_{\zeta}^{\mathbf{B}_n}$ contains the paired force and torque disturbance directions for class $\zeta$.

% For each disturbance class $\zeta \in \{a,s,b\}$, corresponding to arbitrary, swing-induced, and base-motion-induced disturbances, the weakest directional force and torque capacities are defined as

% \begin{subequations}
% \begin{align}
% F_\zeta(\mathcal{C})
% &:=
% \min_{\mathbf{u} \in \mathcal{U}_{\zeta,f}^{\mathbf{B}_n}}
% \max_{\mathbf{f}\in\mathcal{W}_f^{\mathbf{B}_n}(\mathcal{C})}
% \mathbf{f}^{\top}\mathbf{u},
% \label{eq:directional_force_capacity}\\
% T_\zeta(\mathcal{C})
% &:=
% \min_{\mathbf{u} \in \mathcal{U}_{\zeta,\tau}^{\mathbf{B}_n}}
% \max_{\boldsymbol{\tau}\in\mathcal{W}_{\tau}^{\mathbf{B}_n}(\mathcal{C})}
% \boldsymbol{\tau}^{\top}\mathbf{u},
% \label{eq:directional_torque_capacity}
% \end{align}
% \end{subequations}

% where $\mathcal{U}_{\zeta,f}^{\mathbf{B}_n}$ and $\mathcal{U}_{\zeta,\tau}^{\mathbf{B}_n}$ contain the force and torque disturbance directions for class $\zeta$. The capacities are combined as

% \begin{equation}
% s_\zeta(\mathcal{C})
% =
% F_\zeta(\mathcal{C})
% +
% \frac{T_\zeta(\mathcal{C})}{\ell_{\mathrm{nom}}},
% \label{eq:contact_subscore}
% \end{equation}

% where $\ell_{\mathrm{nom}}$ is the nominal distance from the wrench reference point to the active contacts. It is used to express the torque capacity on an equivalent force scale. 

The total contact-configuration score is

\begin{equation}
s(\mathcal{C}) = \sum_\zeta w_\zeta s_\zeta(\mathcal{C}),
\label{eq:contact_config_score}
\end{equation}
where the nonnegative weights $w_\zeta$ specify the relative importance of the disturbance classes and are fixed for all swing-order evaluations.

For each candidate swing order, the scores of the induced contact configurations are sorted in ascending order to form a contact-set score vector. The score vectors are compared lexicographically from their smallest entries, with higher priority given to swing orders with stronger worst-case contact configurations. If the highest-ranked order is kinematically infeasible for the local contact geometry, the next-highest-ranked feasible order is selected. 

\begin{figure*}
    \centering   
    \includegraphics[width=0.95\linewidth]{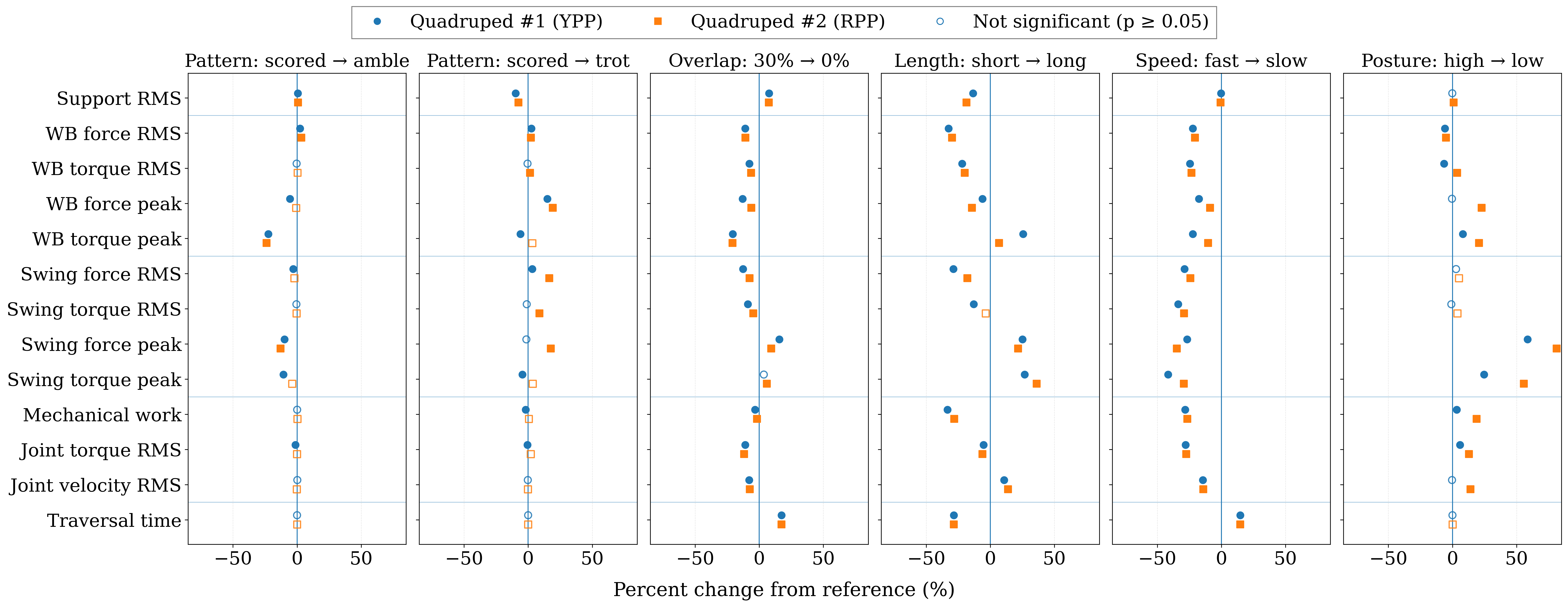}
    \caption{Percentage changes in performance metrics relative to the reference gait for one-at-a-time gait-parameter variations. Higher values are preferred for support capability, whereas lower values are preferred for all other metrics.}
    \vspace{-3mm}
    \label{fig:result_comparison}
\end{figure*}

\subsection{Evaluation Protocol}

Each comparative condition is simulated in 100 randomized environments. The same environment set is used for all parameter conditions and both quadruped morphologies. All trials use identical joint specifications, grasp models, and controller settings.

For each trial, task completion and failure modes, such as kinematic infeasibility, are recorded. Support capability is obtained from the root-mean-square (RMS) value of $F_{\mathrm{support}}(\mathcal{C}(t))$ over the full traversal. For computational tractability, disturbance rejection is evaluated at the eight corners of the robot base along the six positive and negative base frame axis directions. 

Motion-induced disturbance is evaluated from the rate of change of robot momentum. At each time step, the linear momentum rate and the angular momentum rate about the realized base-frame origin are obtained from the simulation and expressed in the base frame to obtain the whole-body motion-induced wrench. The swing-induced wrench is computed analogously from the momentum rates of the links belonging to the limbs currently in swing, with their contributions expressed about the same base-frame origin and aggregated over all swinging limbs. For both wrench measures, the peak and RMS magnitudes of the force and torque components over the full traversal are reported.

Each parameter variant is compared with the reference condition only on environment realizations for which both conditions successfully complete the traversal, yielding paired observations under identical environmental conditions. Statistical significance is assessed for each metric using a paired t-test at the 0.05 significance level.

\begin{table}[t]
    \centering
    \caption{Performance metrics under the reference condition}
    \setlength{\tabcolsep}{3pt}
    \renewcommand{\arraystretch}{1.08}
    \begin{tabular}{l c c}
        \hline
        Metric & Quadruped \#1 & Quadruped \#2 \\
        \hline
        \hline

        \multicolumn{3}{l}{\textit{Support capability}} \\
        $F_{\mathrm{support}}$ RMS [N]
        & 185.2 & 182.1 \\

        \hline
        \multicolumn{3}{l}{\textit{Motion-induced disturbance}} \\
        Whole-body force RMS [N]
        & 24.21 & 19.01 \\

        Whole-body torque RMS [N\,m]
        & 11.55 & 9.66 \\

        Whole-body force peak [N]
        & 79.31 & 63.42 \\

        Whole-body torque peak [N\,m]
        & 37.99 & 32.77 \\

        Swing force RMS [N]
        & 31.83 & 25.36 \\

        Swing torque RMS [N\,m]
        & 35.12 & 27.70 \\

        Swing force peak [N]
        & 212.7 & 212.9 \\

        Swing torque peak [N\,m]
        & 230.0 & 218.7 \\

        \hline
        \multicolumn{3}{l}{\textit{Actuation demand}} \\
        Mechanical work [J]
        & 2154.9 & 2014.0 \\

        Joint torque RMS [N\,m]
        & 3.98 & 3.98 \\

        Joint velocity RMS [rad/s]
        & 0.35 & 0.35 \\

        \hline
        \multicolumn{3}{l}{\textit{Traversal performance}} \\
        Traversal time [s]
        & 103.2 & 103.2 \\

        \hline
    \end{tabular}

    \vspace{0.4em}
    \begin{minipage}{0.97\columnwidth}
        \footnotesize
        \centering
        *Values denote means across successful reference trials.
        % Reference values are reported in absolute units for Quadrupeds
        % \#1/\#2. Variant values indicate percentage changes relative to
        % the corresponding reference condition. A dash indicates no
        % statistically significant difference ($p \geq 0.05$).
        % Arrows indicate the favorable direction of each metric.
    \end{minipage}

    \label{tab:performance_comparison}
    \vspace{-3mm}
\end{table}

\section{Results}
% Over 100 randomized environments, the simulation studies evaluated the effects of swing order, swing-phase overlap, stride length, swing speed, and nominal posture on locomotion performance (i.e., stability and actuation effort) for two quadruped morphologies. 

The simulation studies were used to characterize the parameter-metric relationships within the proposed gait-level framework across randomized environments and two quadruped morphologies. All trials were successfully completed except for one trot trial, four long-stride trials, and six low-posture trials for Quadruped~\#2, which failed due to kinematic infeasibility. Absolute performance metrics under the reference condition are reported in TABLE~\ref{tab:performance_comparison}, while Fig.~\ref{fig:result_comparison} shows the percentage changes produced by each one-at-a-time parameter variation relative to the reference condition.

% The selected values were chosen to produce distinct contact configurations and motion timings while remaining feasible for comparison across both robot morphologies.

% Performance changes relative to the reference condition are summarized in TABLE \ref{tab:performance_comparison}.

\subsection{Swing Order and Phase Overlap Comparison}

% Swing order and phase overlap had the strongest influence on task success. The score-ranked condition succeeded in all 100 trials for both morphologies, whereas canonical swing orders reduced success rates by up to 26\%. Most failures in amble were due to unavailable plans, while trot demonstrated relatively more failures due to limited feasible wrench space, a trend that became stronger at 50\% overlap.

% Across the two morphologies, the metric-level trends were broadly consistent. The score-selected swing order achieved the highest normalized contact score, followed by trot and amble. Amble produced comparatively modest changes, mainly increasing whole-body torque metrics, whereas trot increased whole-body force metrics while reducing peak torque. For overlap variations, the score-selected condition increased disturbance and actuation metrics while reducing traversal time, whereas the 50\% overlap condition reduced contact score, swing-wrench metrics, and mechanical work, with shorter traversal times. These results indicate that score-selected swing order and extended swing phase overlap improved the performance metrics of interest, although greater overlap degraded support capability.

% Swing order primarily affected support capability and motion-induced wrench metrics, whereas phase overlap additionally affected actuation demand and traversal time. 

Compared to the score-ranked reference order, amble reduced peak whole-body torque and peak swing force in both morphologies, while remaining common significant changes were below 4\%. Trot reduced support capability and increased peak whole-body force for both morphologies, with additional increases in the swing-induced wrench metrics for Quadruped~\#2. Eliminating swing-phase overlap increased support capability and reduced the RMS whole-body and swing-induced wrench metrics for both morphologies. Whole-body force and torque peaks decreased, whereas peak swing force increased. Mechanical work, joint-torque RMS, and joint-velocity RMS decreased, while traversal time increased.

\subsection{Stride Length and Swing Speed Comparison}

% A shorter stride length produced the largest increases in disturbance, actuation, and traversal time for both morphologies. In contrast, slower swing speed substantially reduced whole-body and swing-wrench metrics and mechanical work while leaving the normalized contact score nearly unchanged, although traversal time increased. These results suggest that longer stride lengths are generally favorable within kinematic limits, whereas slower speeds may be beneficial when traversal time is not a primary concern.

Longer strides reduced support capability, whole-body force and torque RMS, and swing-force RMS for both morphologies. In contrast, peak swing-induced force and torque increased, while whole-body peak force decreased and peak torque increased. Mechanical work, joint-torque RMS, and traversal time also decreased, whereas joint-velocity RMS increased. Reducing the maximum swing speed decreased both RMS and peak whole-body and swing-induced wrench metrics for both morphologies. Mechanical work, joint-torque RMS, and joint-velocity RMS also decreased, while traversal time increased.

% ; support capability changed by less than 1\% for both morphologies.

\subsection{Nominal Posture and Morphology Comparison}

% The posture variant produced the smallest overall changes. For both morphologies, the normalized contact score decreased, and swing wrench metrics were generally lower than in the reference condition, while changes in whole-body wrench metrics, actuation demand, and traversal time remained minor. Comparing the reference conditions, quadruped \#2 generally showed lower disturbance- and effort-related metrics than quadruped \#1. The majority of whole-body and swing-wrench metrics decreased, along with mechanical work, while swing peak torque and joint torque increased. These findings highlight that favorable nominal postures and robot morphologies can contribute to stable and efficient locomotion.

% The nominal-posture variation produced the smallest overall changes, with no systematic statistically significant effect on the evaluated metrics over the tested range. This contrasts with the clearer effects produced by swing order, phase overlap, stride length, and swing speed.

Lowering the nominal base height increased the peak whole-body and swing wrenches and mechanical work for both morphologies, with substantially larger increases for Quadruped \#2. Under the reference condition, Quadruped~\#1 generally exhibited larger absolute metric values than Quadruped~\#2. These differences reflect morphology-dependent motion realizations. Despite these differences in absolute magnitude, the parameter-induced changes followed similar qualitative trends across the two quadrupeds.

% The nominal-posture variation produced no statistically significant changes in the evaluated metrics for either morphology over the tested range. 

\section{Discussion}

% The simulation study enables systematic examination of the proposed gait-level parameters through their realized effects on locomotion feasibility and performance. Across randomized environments and two quadruped morphologies, the observed trends show how contact sequencing, motion timing, spatial progression, and base–anchor geometry influence the proposed evaluation metrics. While the specific quantitative values depend on the chosen realization, the discussion emphasizes the observed relationships between the gait-level parameters and evaluation metrics.

The simulation study revealed relationships between gait-level parameters and evaluation metrics and trade-offs in locomotion feasibility and performance. Across randomized environments and two quadruped morphologies, similar qualitative trends emerged despite differences in the realized motions and absolute metric values. These relationships can be interpreted by examining how each gait parameter modifies the underlying contact and motion mechanisms.

% For successful trials, variations in gait parameters revealed trade-offs among performance metrics. Support capability was governed by effective support geometry and contact redundancy. The support capability increased under gait parameter settings with larger effective moment arms (e.g., score-ranked swing orders, longer stride lengths, and higher nominal postures) and greater contact redundancy (e.g., lower swing phase overlap). However, support capability was not an isolated performance metric. Larger effective moment arms raised joint torque demand at stance limbs and sensitivity to disturbances, while greater contact redundancy increased traversal time. Support capability was also closely associated with whole-body motion-induced wrench. In the canonical swing order cases, lower support capability was accompanied by a larger whole-body motion-induced wrench, despite comparable swing-induced disturbance relative to the score-ranked condition, suggesting that better-conditioned support sets improve whole-body stability. Thus, parameters that improve support capability should be tuned with careful consideration of their combined effects on joint torques and traversal time.

Support capability was governed by the instantaneous contact configuration, including contact geometry and redundancy. Swing order and stride length reshaped the support geometry, by changing the intermediate contact configurations and their spatial extent, respectively. The score-based swing-order selection, adopted for rapid screening of candidate contact sequences, favored amble-family orders that avoided the reduction in disturbance-rejection capability observed with trot. Longer strides, despite increasing contact spacing, reduced support capability, suggesting that larger moment arms can be offset by an increased imbalance in how individual contact forces contribute to the net moment. This change was accompanied by shorter traversal time, while the long-stride failures indicate that kinematic feasibility can become limiting as the spatial progression increases. Swing-phase overlap affected support through contact redundancy: eliminating overlap removed the two-contact support phases that cannot span the full-dimensional wrench space, thereby increasing support capability at the cost of longer traversal time. These results show that gait parameters affecting support geometry need to be tuned with careful consideration of their combined effects on traversal efficiency and kinematic feasibility.

% Complementary trade-offs were also observed among motion-induced disturbance, actuation demand, and traversal time. Shorter swing phase overlap and slower swing speed generally decreased motion-induced disturbance and actuation demand, but increased traversal time, suggesting that more dynamic locomotion is viable only when support capability is sufficiently well conditioned. By contrast, longer stride length also decreased motion-induced disturbance and actuation demand while shortening traversal time, indicating that longer stride lengths are generally preferable. However, stride length must not be so large as to induce kinematic infeasibility at the extremes of swing or stance. Nominal posture also influenced disturbance by altering the moment arm of swing-induced loading. In short, when tuning gait parameters to reduce motion-induced disturbance and actuation demand, the associated trade-offs in traversal time and kinematic feasibility need to be evaluated holistically.

Complementary trade-offs in motion-induced loading arose through different mechanisms. The reference-condition absolute values show substantially smaller whole-body wrenches than swing-induced wrenches because the whole-body wrench represents the coupled response of the base, stance limbs, and active contacts. Eliminating swing-phase overlap reduced the net whole-body wrench by allowing swing-induced loading to be accommodated through a more redundant anchored configuration. The accompanying reduction in swing-wrench RMS reflects its aggregate definition over all simultaneously swinging limbs, not necessarily a reduction in the intensity of each individual swing. Reducing swing speed acted more directly by lowering the momentum-rate generated by the swinging limbs, thereby reducing both RMS and peak motion-induced loading together with actuation demand, at the cost of increased traversal time. Longer strides reduced the number of contact transitions and the acceleration changes during swing, approach, and retraction, contributing to lower RMS loading and mechanical work. In contrast, the larger excursions brought the limbs closer to kinematically ill-conditioned configurations, consistent with the increased joint velocity and transient swing-induced peaks.

Another parameter with a substantial effect on kinematic realization was base height, which changes the relative geometry between the base and the available anchors. Unlike terrestrial locomotion, where this height directly affects gravitational potential energy and COM dynamics, its role in microgravity is primarily geometric. Lowering the base height required the limbs to operate in less favorable kinematic configurations, producing larger swing-induced wrenches, as in the longer-stride case.

% Although the two morphologies exhibited different absolute metric values, they showed similar qualitative responses to variations in the gait parameters. This suggests that the dominant trends were primarily associated with gait-level design choices rather than the specific morphology realization.

% Although changing this geometry can alter the resulting joint trajectories and kinematic realization, no systematic changes in the evaluated metrics were observed over the tested height range. 

% Overall, the observed trends reveal coupled trade-offs among contact support, motion-induced loading, and traversal performance. Across the tested conditions, favorable performance was associated with contact configurations that maintained robust support while limiting motion-induced wrench. These findings suggest that effective gait-level design for microgravity locomotion requires coordinated selection of motion parameters to balance these objectives. In addition, the similar parameter–metric relationships observed across the two quadruped morphologies further indicate that the framework is not specific to a single robot realization.

Although no individual parameter variation consistently improved all the evaluation metrics considered, some conditions resulted in more favorable overall outcomes under the investigated settings. The selected gait pattern outperformed trot, while its performance was not significantly different from amble. Removing swing-phase overlap and increasing stride length were generally beneficial, although both involved trade-offs in motion-induced peak wrench, and removing swing-phase overlap additionally increased traversal time. Similarly, a slower travel speed reduced motion-induced wrenches and actuation demands at the cost of longer traversal time. For the selected morphologies, a higher base height was preferable to avoid kinematically ill-conditioned states.

Overall, the results demonstrate that gait-level parameter variations, including changes in contact organization and motion execution, can affect locomotion feasibility and performance through coupled changes in support capability, motion-induced loading, actuation demand, and traversal time. Despite morphology-dependent differences in the realized motions and absolute metric values, the qualitative parameter-metric relationships remained broadly consistent across the two tested quadrupeds. This suggests that the dominant trends were primarily associated with gait-level design choices rather than the specific morphology realization, supporting the proposed gait-level parameterization as a common and interpretable basis for comparing locomotion strategies across robot realizations.

More broadly, the identified parameter-metric relationships can inform parameter selection according to robot capabilities, environmental conditions, and mission objectives. Future work can integrate the proposed parameters and evaluation metrics with model-based or learning-based methods for condition-dependent parameter selection, while preserving an interpretable connection between gait-level design choices and locomotion outcomes.

% This study uses a quadruped platform to investigate the parameters governing dynamic locomotion performance in microgravity. While the current validation is limited to quadrupeds, the framework is extendable to other limbed morphologies, such as bipeds and hexapods, with corresponding trade-offs in contact redundancy, dynamic feasibility, and kinematic complexity. The study also adopts simplifying assumptions, including prior environment knowledge and abstracted grasping models, to isolate gait-parameter effects from perception and grasp uncertainty. Future work will relax these assumptions through SLAM-based environment estimation, probabilistic grasp-success modeling, and more sophisticated gripper-specific contact models.

\section{Conclusion}

This study presents a gait-level framework for analyzing grasp-based dynamic locomotion in microgravity under coupled feasibility and performance considerations. The framework formulates the relevant locomotion objectives and relates gait-level motion parameters to quantitative evaluation metrics. Across randomized environments and two quadruped morphologies, the simulations revealed broadly consistent parameter–metric relationships despite morphology-dependent differences in absolute performance. These relationships exposed interpretable trade-offs among contact support, motion-induced loading, kinematic feasibility, actuation demand, and traversal time. The proposed framework therefore provides a systematic basis for comparing gait-level motion designs with respect to locomotion constraints and objectives. Future work will extend this framework toward condition-dependent parameter selection.

\bibliographystyle{IEEEtran}
\bibliography{reference}

\end{document}